\documentclass[runningheads]{llncs}
\usepackage[T1]{fontenc}
\usepackage{graphicx}
\usepackage{amsmath}
\usepackage{amssymb}
\usepackage{layout}
\usepackage{multirow}
\usepackage{adjustbox}
\usepackage{float}
\usepackage{tabularray}
\usepackage{hyperref}

\begin{document}
\title{Controllable 3D object Generation\\with Single Image Prompt}

\author{Jaeseok Lee\inst{1} \and
Jaekoo Lee\inst{1}$^*$}

\authorrunning{Lee et al.}

\institute{College of Computer Science, Kookmin University, Seoul, Korea\\
\textsuperscript{*}Correspondence: jaekoo@kookmin.ac.kr}

\maketitle  

\begin{abstract}
Recently, the impressive generative capabilities of diffusion models have been demonstrated, producing images with remarkable fidelity. Particularly, existing methods for the 3D object generation tasks, which is one of the fastest-growing segments in computer vision,     predominantly use text-to-image diffusion models with textual inversion which train a pseudo text prompt to describe the given image. In practice, various text-to-image generative models employ textual inversion to learn concepts or styles of target object in the pseudo text prompt embedding space, thereby generating sophisticated outputs.
However, textual inversion requires additional training time and lacks control ability. To tackle this issues, we propose two innovative methods: (1) using an off-the-shelf image adapter that generates 3D objects without textual inversion, offering enhanced control over conditions such as depth, pose, and text. (2) a depth conditioned warmup strategy to enhance 3D consistency. In experimental results, ours show qualitatively and quantitatively comparable performance and improved 3D consistency to the existing text-inversion-based alternatives. Furthermore, we conduct a user study to assess (i) how well results match the input image and (ii) whether 3D consistency is maintained. User study results show that our model outperforms the alternatives, validating the effectiveness of our approaches. Our code is available at GitHub repository:\url{https://github.com/Seooooooogi/Control3D_IP/}

\end{abstract}

\begin{figure}[!ht]
    \centering
    \includegraphics[width=1.0\linewidth]{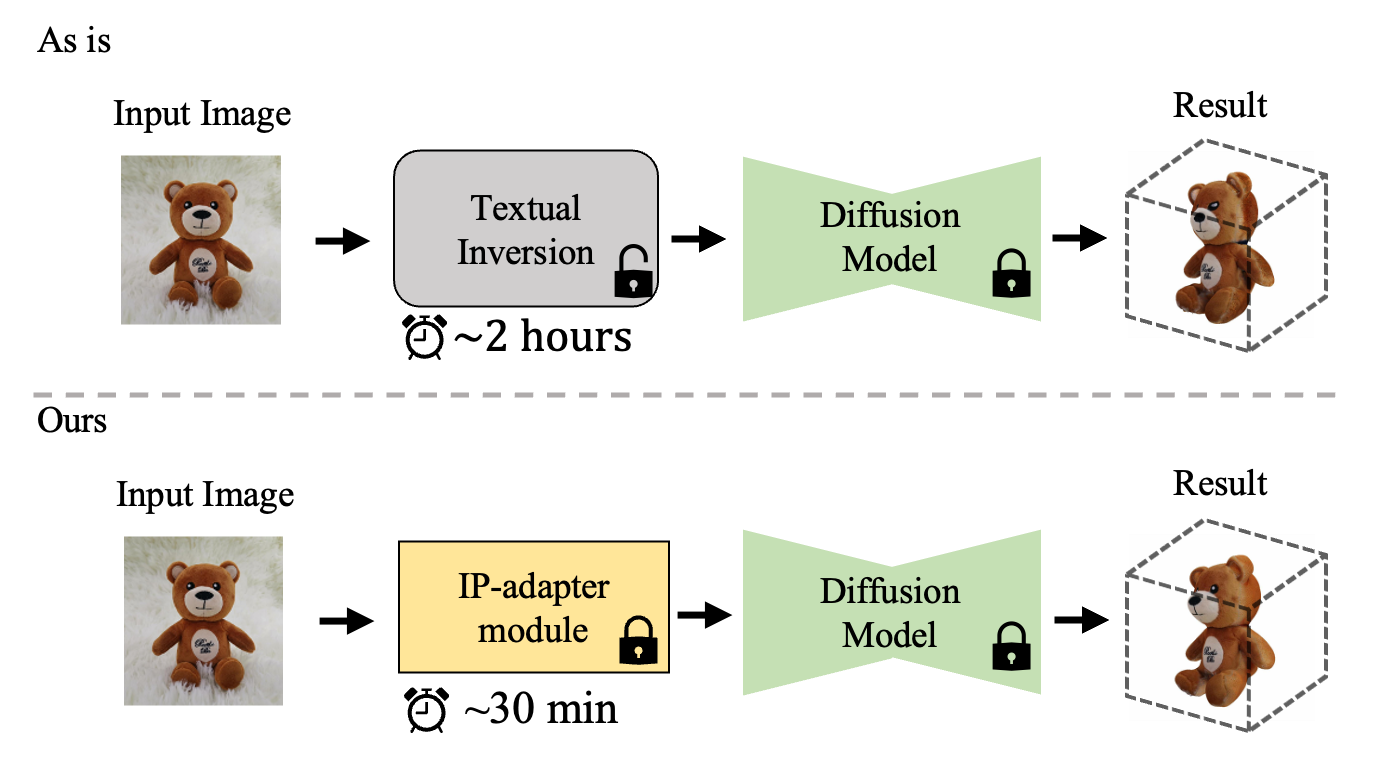}
    \caption{Comparison between (a) existing method using text-guided priors via textual inversion and (b) our method that employs a single image prompt.}
    \label{fig:introduction}
\end{figure}

\section{Introduction}
Generative models are widely used in real-world industrial applications, including creating artworks~\cite{Rombach_2022_CVPR}, video generation~\cite{singer2022makeavideo,sora}, automatic colorization~\cite{lee2022bridging,zabari2023diffusing}, virtual try-on~\cite{Zhu_2023_CVPR_tryondiffusion}, and 3D computer graphics~\cite{poole2022dreamfusion,melaskyriazi2023realfusion,Mohammad_Khalid_2022,chen2023text2tex,qian2023magic123,Chen_2023_ICCV}. Especially, 3D object generation tasks have significant demand in computer vision industries such as Artificial Reality (AR), Virtual Reality (VR), and Gaming. Conventionally, generating 3D objects has been a delicate and intricate process that requires experts well-versed in 3D graphics tools with a considerable amount of time. However, recent breakthroughs have introduced 3D generative models (e.g., text-to-3D model and image-to-3D model) that allow even non-experts to effortlessly produce 3D objects.

Especially, several methods~\cite{melaskyriazi2023realfusion,qian2023magic123} have achieved significant advancements in image-to-3D generation by using text-guided priors and textual inversion~\cite{gal2022image}, which is subsequently used in text-to-image diffusion model. However, as shown in Fig.~\ref{fig:introduction}, textual inversion requires additional time and cost to get a new prompt because it needs training process to optimize pseudo text prompt.

To address these issues, we propose the following contributions.
\begin{itemize}
    \item We propose a controllable image prompt score distillation sampling(called SDS) method that uses a single image as a prompt without textual inversion to generate novel views of 3D object.
    \item Our method leverages depth estimated from the 3D object as a condition for ControlNet~\cite{zhang2023adding} to prevent the degradation of 3D consistency, and employs a depth conditioned warmup strategy to alleviate instability of depth in early training epochs.
    \item We experimentally show that our proposed method generates diverse, controllable 3D objects using a given single image and optional prompts such as depth, pose, and text. Our method quantitatively and qualitatively outperforms alternative approaches.
    \item We also conduct a user study that participants score two evaluation metrics: fidelity and 3D consistency. Our method exhibits superior performance compared to SOTA methods.
\end{itemize}
\section{Related Work}

\textbf{Neural 3D Representations.} Neural Radiance Fields (NeRF)~\cite{mildenhall2020nerf} trains deep neural network with sparse input views and camera poses to optimize the inherent continuous volumetric scene function, synthesizing novel views of given scenes. Although NeRF has capability of 3D reconstruction task, optimization process is slow and hard to get high-resolution results. To overcome computational cost, Instant-NGP~\cite{mueller2022instant} uses smaller neural networks, leverages multi resolution hash grids without sacrificing quality. On the other hand, Deep Marching Tetrahedra (DMTet)~\cite{shen2021dmtet} leverages novel hybrid implicit-explicit 3D representations, achieving high resolution, finer geometric details with fewer artifacts with efficient memory consumption. In here, we utilize Instant-NGP and DMTet for generate 3D object from a single image, aiming for higher quality results while maintaining memory efficiency.

\textbf{Diffusion Models.} In computer vision tasks such as inpainting~\cite{lugmayr2022repaint,saharia2022palette}, image editing~\cite{meng2022sdedit,sinha2021d2c}, super-resolution~\cite{li2021srdiff,saharia2021image}, image generation~\cite{ramesh2022hierarchical,Rombach_2022_CVPR} and video generation~\cite{singer2022makeavideo,sora}, diffusion models are gaining popularity due to great result quality. Diffusion models train neural networks to denoise images blurred with Gaussian noise and to reverse the diffusion process~\cite{ho2020denoising}. Especially, Stable diffusion~\cite{rombach2022highresolution}, which is open source, achieved great success with datasets consisting of large scale text-image pairs~\cite{schuhmann2022laion}.
Subsequent work, ControlNet, leverages the expressiveness of the Stable diffusion model, expanding it to incorporate a variety of conditional conditional priors such as sketches, depth, and human poses~\cite{zhang2023adding}.
Gal, Rinon~\textit{et al.}~\cite{gal2022image} introduce textual inversion, which optimizes a pseudo text prompt by using only 3-5 images of a user-provided concept to utilize the expressiveness of Stable diffusion model. However, it requires additional training time and usually needs several images. To overcome these shortcomings, IP-Adapter~\cite{ye2023ip-adapter} proposes a lightweight adapter to directly input an image prompt into a pretrained text-to-image diffusion model, thereby facilitating multi-modal image generation without additional training time.

\textbf{3D Generative Models.} By integrating NeRF and diffusion models, several approaches aim to train 3D generative models, either by using text~\cite{poole2022dreamfusion,lin2023magic3d,Chen_2023_ICCV} or a single image~\cite{melaskyriazi2023realfusion,qian2023magic123}. For text-to-3D generation, DreamFusion~\cite{poole2022dreamfusion} proposes score distillation sampling to guide the training of NeRF models using a pretrained text-to-image diffusion model. Although DreamFusion successfully achieved first text-to-3D generative task with pretrained diffusion models, it fails to generate high-resolution objects, requiring up to one day to train a single 3D object. To produce high-resolution 3D objects, several methods are proposed. Magic3D~\cite{lin2023magic3d} integrates DMTet~\cite{shen2021dmtet} during the training process. Fantasia3D~\cite{Chen_2023_ICCV} disentangles geometry and appearance to generate high-fidelity 3D objects that closely align with real graphics rendering. 
For image-to-3D generation task, RealFusion~\cite{melaskyriazi2023realfusion} employs textual inversion from an image to derive custom tokens for training a 3D generative model instead of text prompts. On the other hand, Zero-1-to-3~\cite{liu2023zero1to3} finetunes the Stable diffusion model to simultaneously input an image and geometry-related camera pose priors, facilitating the synthesis of images from specific 3D viewpoints. Magic123~\cite{qian2023magic123} combines the Stable diffusion model with textual inversion and Zero-1-to-3 simultaneously. Despite RealFusion and Magic123 utilize textual inversion for single image, these methods often fail to capture fine details of given images and requires additional training time.

\begin{figure}[!t]
    \centering
    \begin{adjustbox}{width=1.4\linewidth, center}
        \includegraphics{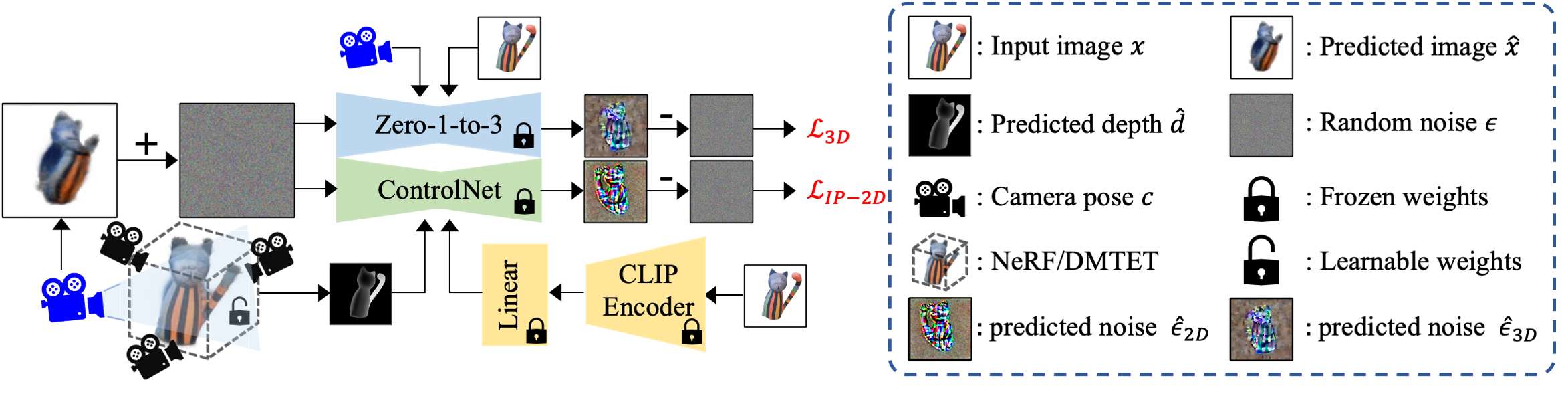}
    \end{adjustbox}
    \caption{Overall architecture of our proposed 3D generative model. During training, our method iteratively use the following stages: (i) controllable image prompt score distillation sampling, and (ii) depth conditioned warmup strategy.}
    \label{fig:overview}
\end{figure}

\section{Methods}
The main goal of our method is to synthesize a 3D object without relying on text prompts obtained by textual inversion, instead directly using controllable image prompts (e.g., only image or with optional conditions). To achieve this goal, as shown in Fig.~\ref{fig:overview}, we introduce \textit{controllable image prompt score distillation sampling} and \textit{depth conditioned warmup strategy}.
We build the proposed model, which consists of two main components, using the Magic123~\cite{qian2023magic123} architecture as the baseline. The proposed model performs alternately two stage training process (i.e., coarse-to-fine training). During the 1st stage for training coarse grained representation, we train NeRF. In detail, initialized NeRF model with learnable parameters $\theta$, predicts the volume density $\sigma$ and color $c$ of each pixel given a random camera pose $c$. 

\begin{equation}
    (c, \sigma) = NeRF(\theta; c)
\end{equation}
By computing each pixel's volume density and color, NeRF predicts image $\hat{x}\in\mathbb{R}^{H\times W \times 3}$ and depth $\hat{d}\in\mathbb{R}^{H\times W \times 1}$. 

When the 1st training stage finished, we finetune the NeRF with DMTet to represent the target 3D object at high resolution during 2nd stage for training fine grained representation. Unlike NeRF which has entangled geometry and texture representations, DMTet has disentangled geometry and texture representations. 
For geometry representation, DMTet uses a deformable tetrahydral grid $(V_{T}, T)$~\cite{shen2021dmtet}, where tetrahydral grid denotes $T$ and its vertices denote $V_T$. DMTet represents the 3D object using a Signed Distance Function  (SDF) $s$ at vertex $v_{i}\in V_{T}$ and a triangle deformation vector $\Delta v_i$. In here, $s$ is initialized from the NeRF, while triangle deformation vector $\Delta v_i$ is initialized as zero. For texture representation, DMTet uses a neural color field, as mentioned in Magic3D~\cite{lin2023magic3d}, is also initialized from the NeRF.

\begin{figure}[!t]
    \begin{adjustbox}{width=1.2\linewidth, center}
        \includegraphics{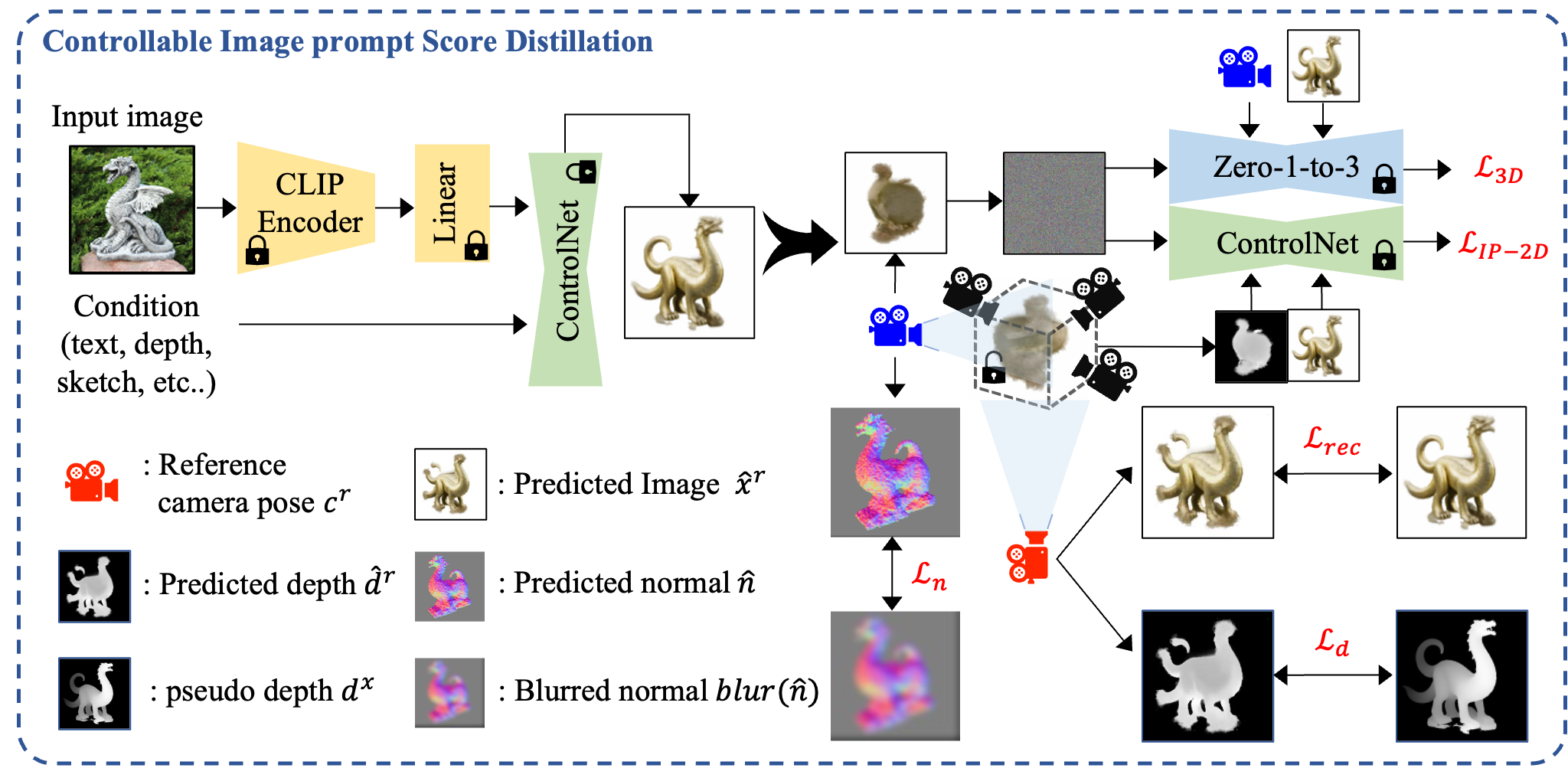} 
    \end{adjustbox}
    \caption{An overview of controllable image prompt score distillation sampling. See details on Sec~\ref{sec1}.}
    \label{fig:IPSDS}
\end{figure}

\subsection{Controllable Image Prompt Score Distillation Sampling}\label{sec1}
As shown in Fig.~\ref{fig:IPSDS}, a single input image $x$, optionally including various conditions such as text, depth, sketch, is embedded into a controllable image prompt through the well-known pretrained Image Prompt Adapter (IP-Adapter)~\cite{ye2023ip-adapter}, which consists of the pretrained CLIP~\cite{radford2021learning} image encoder and the single linear layer. The embedded controllable image prompt and depth $\hat{d}$, obtained from NeRF/DMTet, are then injected as conditions into a stable diffusion model like ControlNet~\cite{zhang2023adding} to predict $\hat{\epsilon}_{2D}$ .

The $\hat{x}$ obtained from NeRF/DMTet is added time-dependent noise $\epsilon$ to produce a noisy latent $\hat{x}_t$. $\hat{x}_t$ is then fed into the ControlNet\cite{zhang2023adding}. Consequently, ControlNet outputs $\hat{\epsilon}_{2D}$, corresponding to the random diffusion timestep $t$. The loss function $\mathcal{L}_{IP-2D}$ for 2D score distillation sampling is as follows:

\begin{figure}[!t]
    \begin{adjustbox}{width=1.3\linewidth, center}
        \includegraphics{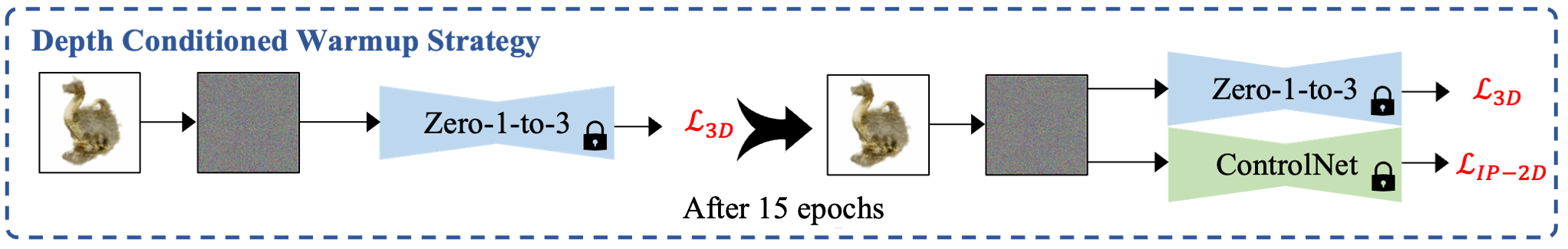}
    \end{adjustbox}
    \caption{An overview of depth conditioned warmup strategy. See details on Sec~\ref{sec2}.}
    \label{fig:DCWS}
\end{figure}

\begin{equation}
    \mathcal{L}_{IP-2D} = \mathbb{E}_{t,\epsilon}\Big[w(t)(\hat{\epsilon}_{2D}(\hat{z}_t;x,\hat{d},t) - \epsilon)\frac{\partial x}{\partial \theta}\Big]
\end{equation}
where $w(t)$ represents the time-dependent weighting function. By directly injecting input image into the stable diffusion model, the 2D score distillation sampling loss guides NeRF/DMTet to adhere to the input image from any camera pose.
 
Similar to ControlNet, Zero-1-to-3 predicts $\hat{\epsilon}_{3D}$ when given $\hat{x}_t$ as input. The loss function $\mathcal{L}_{3D}$ for 3D score distillation sampling is as follows:
\begin{equation}
    \mathcal{L}_{3D} = \mathbb{E}_{t,\epsilon}\Big[w(t)(\hat{\epsilon}_{3D}(\hat{z}_t;x,c,t) - \epsilon)\frac{\partial x}{\partial \theta}\Big]
\end{equation}

\noindent{Although $\mathcal{L}_{3D}$ serves the same purpose as $\mathcal{L}_{IP-2D}$, Magic123~\cite{qian2023magic123} found that training with $\mathcal{L}_{3D}$ can lead NeRF/DMTet to have better 3D consistency.}

Additionally, a primary requirement of image-to-3D generative models is to reconstruct input image from a reference camera pose $c^{r}$. Given the input image $x$ masked with $\mathcal{M}$, which is acquired by a dense prediction transformer~\cite{Ranftl2021}, and the predicted image $\hat{x^{r}}$ masked with $\mathcal{M}^{r}$, we perform the reconstruction by comparing $x$ and $\hat{x^{r}}$. The reconstruction loss $\mathcal{L}_{rec}$ is as follows:

\begin{equation}
    \mathcal{L}_{rec} = \| \mathcal{M} \odot (x - \hat{x}^{r})\|_{2}^{2} + \| \mathcal{M} - \mathcal{M}^{r}\|_{2}^{2}
\end{equation}
\noindent{where $\odot$ is element-wise product.}
Xu, Dejia~\textit{et al.}~\cite{xu2023neurallift} introduce that reconstructing 3D object from 2D images often results in flat geometry. To alleviate this issue, we incorporate a monocular depth regularization loss $\mathcal{L}_{d}$, inspired by Magic123~\cite{qian2023magic123}. Given predicted depth $\hat{d}^{r}$ from the reference camera pose $c^{r}$ and the input image's pseudo depth $d^{x}$, estimated by pretrained monocular depth estimator~\cite{ranftl2020towards}, the loss function $\mathcal{L}_{d}$ is calculated using the negative Pearson correlation between $\hat{d}^{r}$ and ${d}^{x}$, where are masked with $\mathcal{M}$.

\begin{equation}
    \mathcal{L}_{d} = \frac{1}{2}\left[1 - \frac{Cov(d^{x}\odot \mathcal{M}, \hat{d}^{r}\odot\mathcal{M})}{\sigma(d^{x}\odot \mathcal{M}), \sigma(\hat{d}^{r}\odot \mathcal{M})} \right]
\end{equation}

The normal smoothness loss function $\mathcal{L}_{n}$ is designed to ensure smooth normals for the 3D object, as generated 3D objects often exhibit noisy artifacts on their surfaces. Specifically, following RealFusion~\cite{melaskyriazi2023realfusion}, we compare the normals $\hat{n}\in \mathbb{R}^{ H\times W \times 3}$ both before and after applying a Gaussian filter $blur$. Then the loss function $\mathcal{L}_{n}$ is defined as follows:

\begin{equation}
    \mathcal{L}_{n} =  \left \|  \hat{n} - stopgrad(blur(\hat{n}, k))\right \|_{2}^{2}
\end{equation}

\noindent{where $stopgrad$ is stop-gradient operation. We use a kernel size $k=9$ for the Gaussian filter.}

Ultimately, we use the following total loss $\mathcal{L}_{total}$ as follows: 
\begin{equation}
\mathcal{L}_{total} = \lambda_{IP-2D}\mathcal{L}_{IP-2D} + \lambda_{3D}\mathcal{L}_{3D}+\lambda_{d}\mathcal{L}_{d} + \lambda_{n}\mathcal{L}_{n} + \mathcal{L}_{rec}
\end{equation}

\begin{figure}
    \begin{adjustbox}{width=1.2\linewidth, center}
        \includegraphics{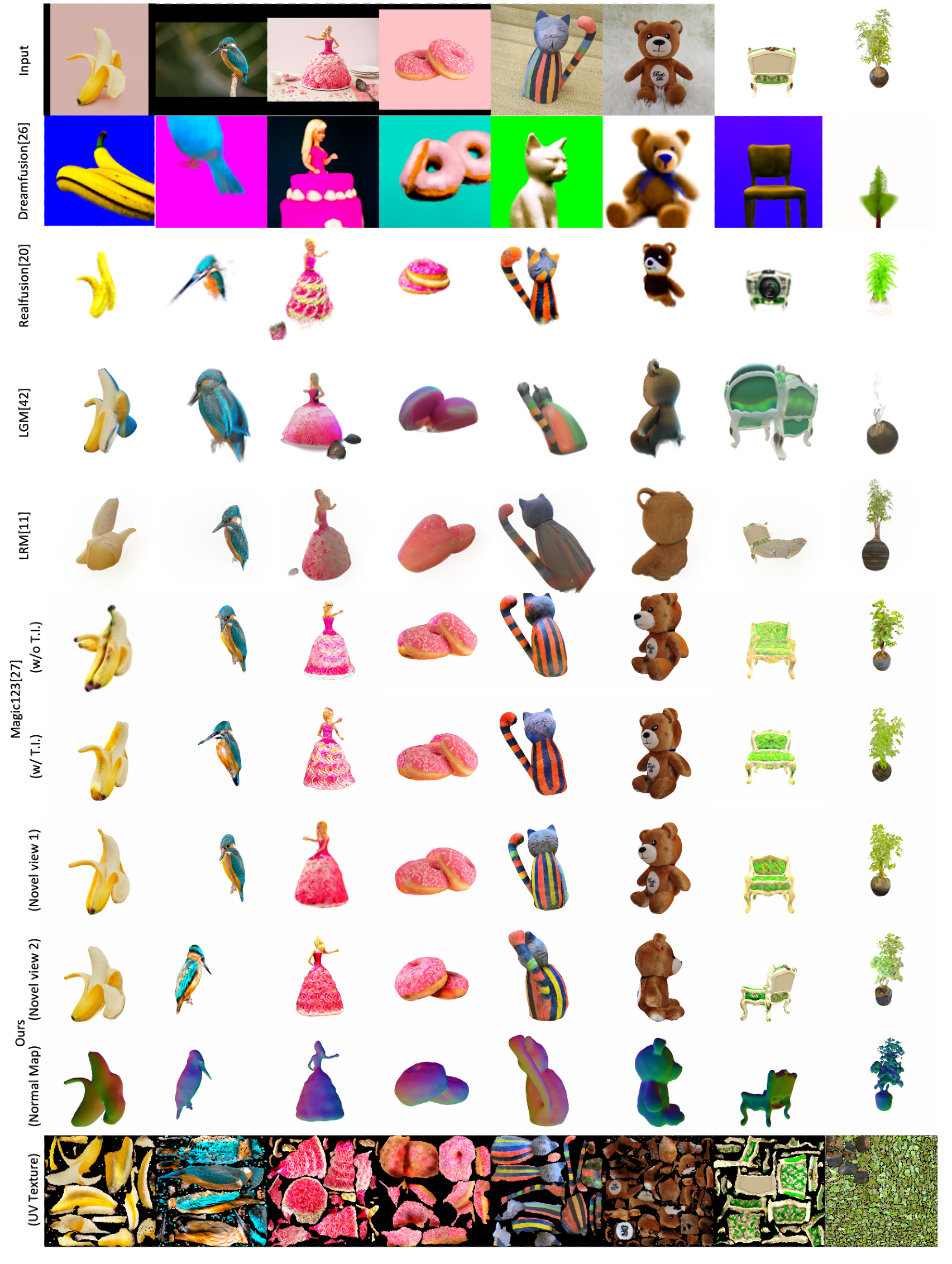}
    \end{adjustbox}
    \caption{Qualitative image-to-3D generation performance comparison with SOTA alternatives (Dreamfusion~\cite{poole2022dreamfusion}, RealFusion~\cite{melaskyriazi2023realfusion}, LGM~\cite{tang2024lgm}, LRM~\cite{hong2023lrm}}, Zero-1-to-3~\cite{liu2023zero1to3}, Magic123~\cite{qian2023magic123}) on realfusion15~\cite{melaskyriazi2023realfusion} and NeRF4~\cite{mildenhall2020nerf} datasets. More detailed results can be found on our GitHub.
    \label{fig:qualitative}
\end{figure}

\begin{figure}
    \begin{adjustbox}{width=1.2\linewidth, center}
        \includegraphics{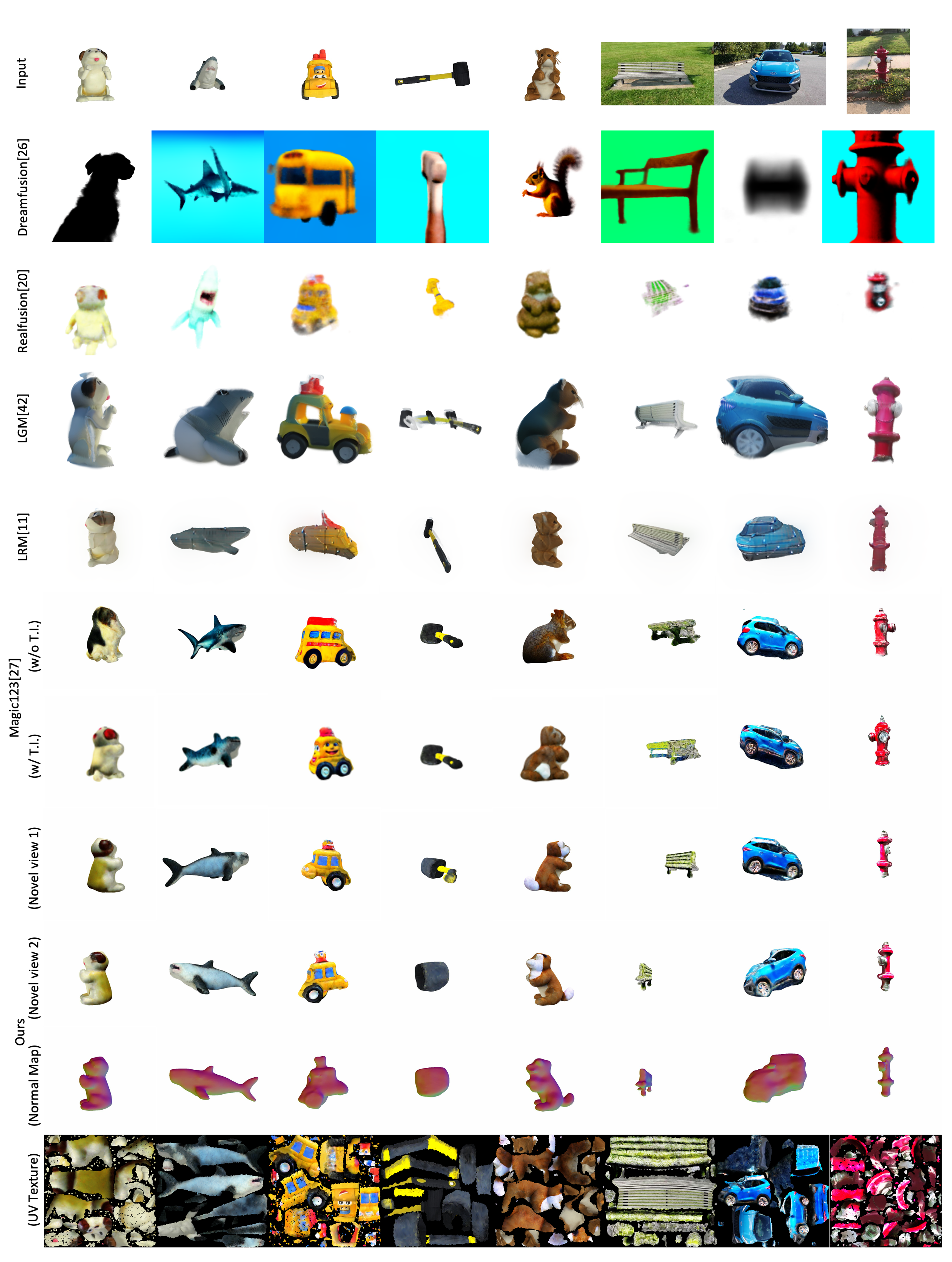}
    \end{adjustbox}
    \caption{Comparison of qualitative image-to-3D generation performance with the previous methods (Dreamfusion~\cite{poole2022dreamfusion}, RealFusion~\cite{melaskyriazi2023realfusion}, LGM~\cite{tang2024lgm}, LRM~\cite{hong2023lrm}, Zero-1-to-3~\cite{liu2023zero1to3}, Magic123~\cite{qian2023magic123}) on GSO~\cite{downs2022googlescannedobjectshighquality} and CO3D~\cite{reizenstein21co3d} datasets. More detailed results can be found on our GitHub.}
    \label{fig:qualitative_2}
\end{figure}


\subsection{Depth Conditioned Warmup Strategy}\label{sec2}
Typically, text-to-image diffusion models suffer from 3D inconsistencies due to insufficient 3D field information~\cite{poole2022dreamfusion}. To mitigate this issue, we integrate the depth $\hat{d}$ obtained from NeRF/DMTet model as a condition for the ControlNet.
As shown in Fig.~\ref{fig:DCWS}, ControlNet~\cite{zhang2023adding} injecting depth $\hat{d}$ introduce a 3D prior into the NeRF/DMTet, thereby enhancing 3D consistency. However, $\hat{d}$ tends to be unstable in early epochs. Therefore, for the initial 15 epochs, we only use the Zero-1-to-3, exploiting on its enhanced geometric capabilities during the 1st stage of training. After 1st stage, we use both ControlNet and Zero-1-to-3 for 2nd stage of training.

Consequently, depth conditioned warmup strategy optimizes a unified 3D field, ensuring that the image synthesized from any viewpoint aligns with high probabilities as evaluated by Zero-1-to-3 by distilling pretrained diffusion models with a depth condition to refine the 3D consistency.


\begin{figure}[!t]
    \centering
    \includegraphics[width=1.0\linewidth]{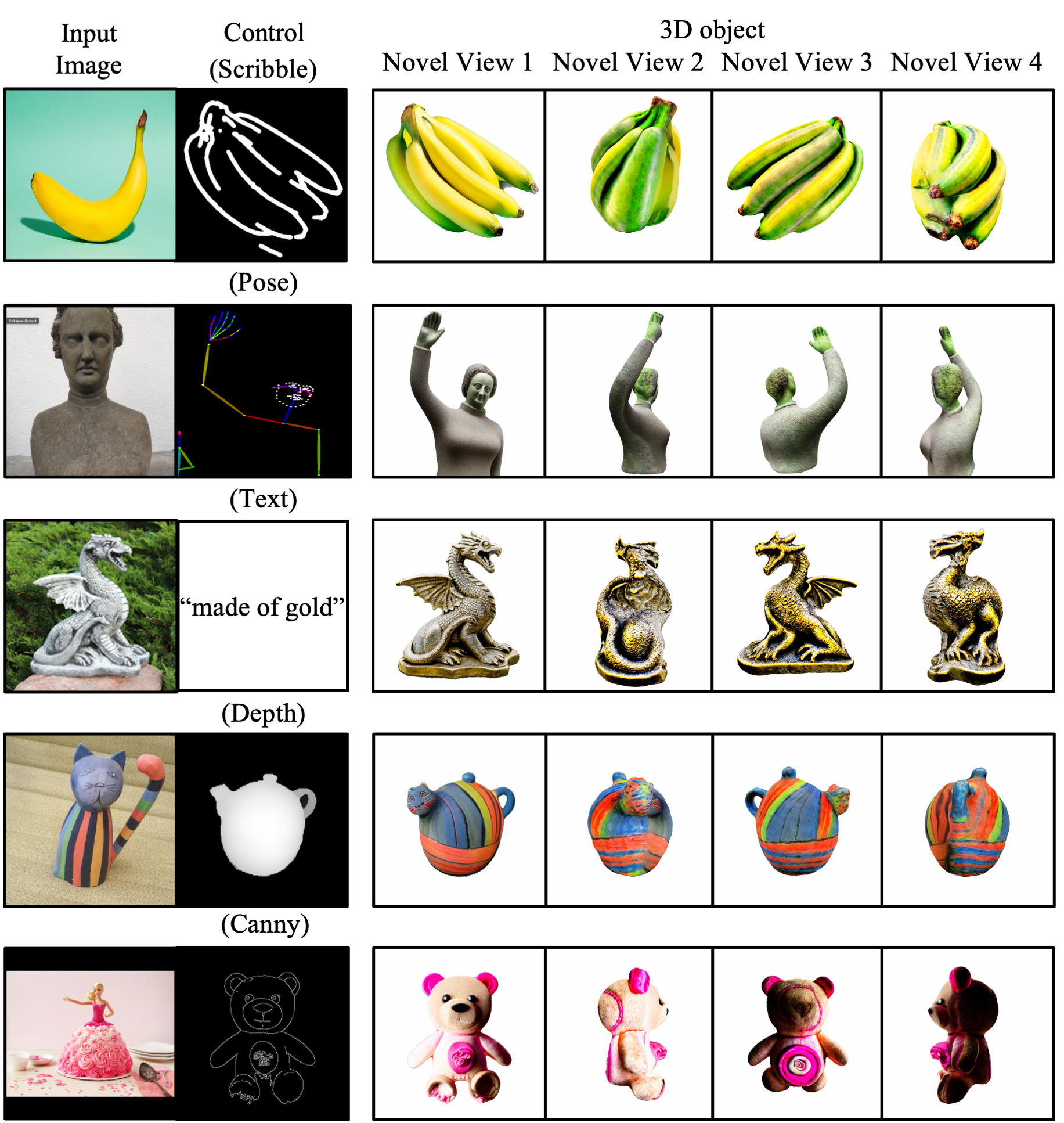}
    \caption{Various controllable 3D synthesis with image prompt and optional conditions.}
    \label{fig:control}
\end{figure}
\section{Experiments}
\subsection{Implementation Details} 
We use Instant-NGP~\cite{mueller2022instant} as the NeRF backbone model, consisting of three layers with 64 hidden dimensions, which is same as~\cite{poole2022dreamfusion,qian2023magic123}. We integrates depth-conditioned ControlNet v1.1 with Stable Diffusion v1.5 and the pretrained Zero-1-to-3 model ($100,500$ iteration checkpoint). For the IP-Adapter, we utilize a vanilla model which extracts 4 tokens from the input image. We maintain the same hyperparameters, including the elevation angle for the input image, for all experiments.

\textbf{Datasets.} We conduct experiments on the publicly available NeRF4~\cite{mildenhall2020nerf}, Realfusion15 (RF15)~\cite{melaskyriazi2023realfusion} , Google Scanned Objects (GSO)~\cite{downs2022googlescannedobjectshighquality}, and Common Objects in 3D (CO3D)~\cite{reizenstein21co3d} datasets. The NeRF4 dataset includes one simple object (mic), two objects with complex geometries (ficus, drums), and one challenging case (chair) where the input image is from the back view. The RealFusion15 dataset comprises real-world photos, including bananas, Barbie cake, bird sparrow, blue bird, cactus, cat statue, colorful teapot, fish, metal dragon statue, microphone, stone dragon statue, teddy bear, two cherries, two donuts, and watercolor horse. 
The GSO dataset includes 3D scans of common household items, providing five views for each object. We used single images of five objects: squirrel, school bus, hammer, dog, and shark. The CO3D dataset offers multi-view images of real-world objects from 50 MS-COCO~\cite{lin2015microsoftcococommonobjects} categories in the form of video frames. For our experiments, we used four scenes: bench, car, bicycle, and hydrant.

\subsection{Quantitative Results}
We perform quantitative experiments using contrastive language-image pre-training-similarity (CLIP-similarity)~\cite{radford2021learning,melaskyriazi2023realfusion} and adjacent learned perceptual image patch similarity (A-LPIPS)~\cite{hong2023debiasing} as evaluation metrics for 3D object generation. CLIP-similarity measures the consistency between the rendered image from evenly spaced azimuths and the category of its corresponding reference image. A-LPIPS is derived from the learned perceptual image patch similarity (LPIPS) metric~\cite{zhang2018unreasonable}, predicted on the notion that two images from adjacent viewpoints should be perceptually similar if the the 3D object is consistent. To evaluate 3D consistency, we measure the average A-LPIPS between images adjancent in viewpoint, maintaining the same elevation angle as used in the CLIP-similarity assessment. For A-LPIPS measurements, we employ neural network backbones such as Alex~\cite{krizhevsky2012imagenet}, and VGG~\cite{simonyan2015deep}, as referenced in previous work~\cite{hong2023debiasing}.

As shown in Table~\ref{table:1}, quantitative experiments compare our method with the state-of-the-art (SOTA) alternatives described in~\cite{qian2023magic123,hong2023lrm,tang2024lgm}. We verify the superiority of our method, which relies solely on image prompts without textual inversion. In practice, the baseline having textual inversion needs to learn a pseudo text prompt from the input image, which requires an additional 1-2 hours of training time. In contrast, our method directly utilizes the input image, significantly reducing training time.

In the CLIP-similarity metric, our method leverages a depth-conditioned warmup strategy to preserve geometry during the training of the target 3D object, demonstrating improved 3D consistency compared to alternatives. While LRM and LGM perform well in terms of the A-LPIPS metric, our approach excels in CLIP-similarity, indicating that our method generates 3D objects that more accurately align with the input image.

However, these evaluation metrics often fail to capture visual realism on 3D objects. To address the limitations of quantitative evaluations, we further evaluate and visualize ours using comparison from novel view synthesis.

\begin{figure}[!t]
    \centering
    \includegraphics[width=0.8\linewidth]{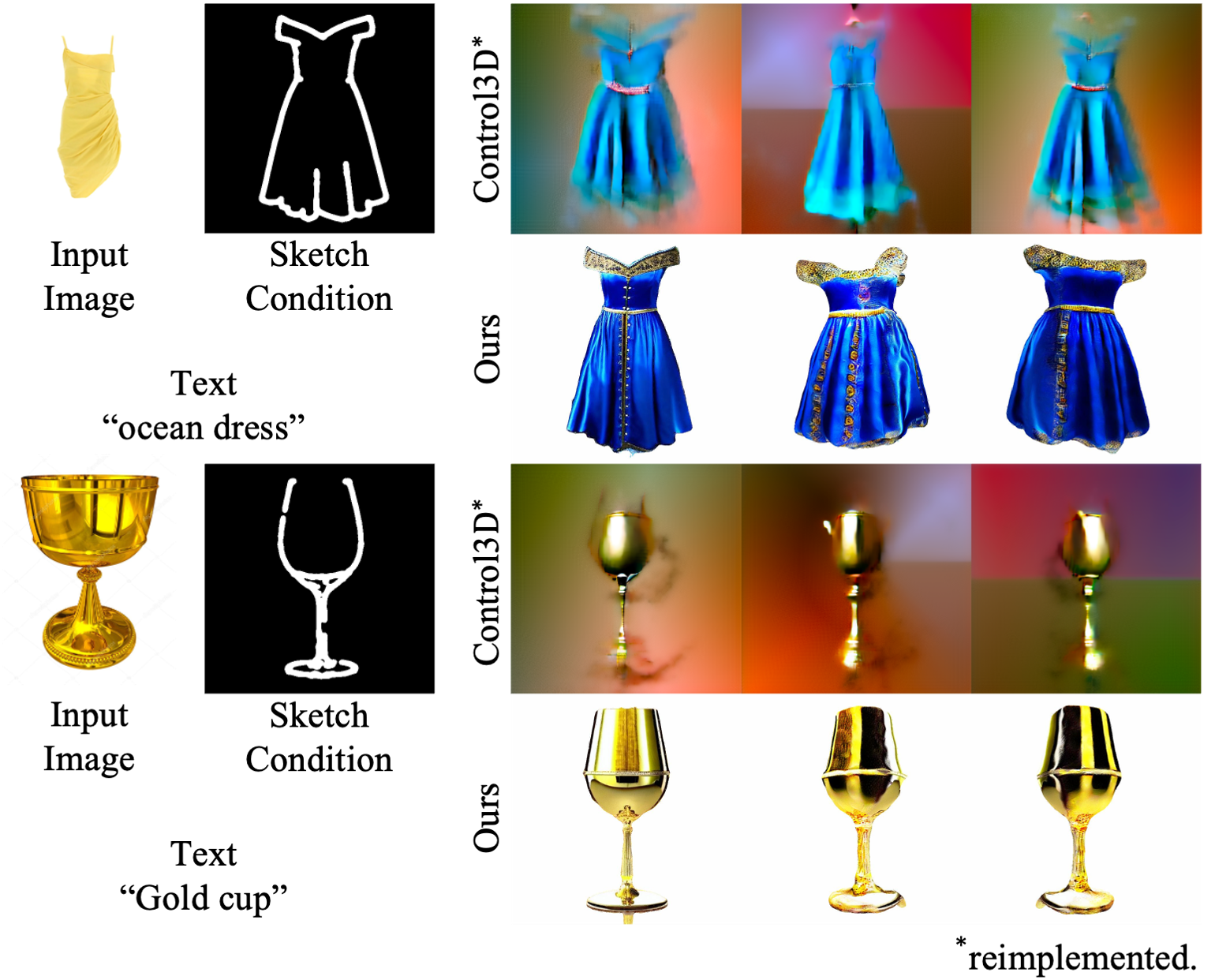}
    \caption{Qualitative comparison with Control3D~\cite{chen2023control3d} and ours on sketch condition. Note that Control3D takes sketch and text as input, not image.}
    \label{fig:control_comparison}
\end{figure}

\begin{figure}[!t]
    \centering
    \begin{adjustbox}{width=1.0\linewidth, center}
        \includegraphics{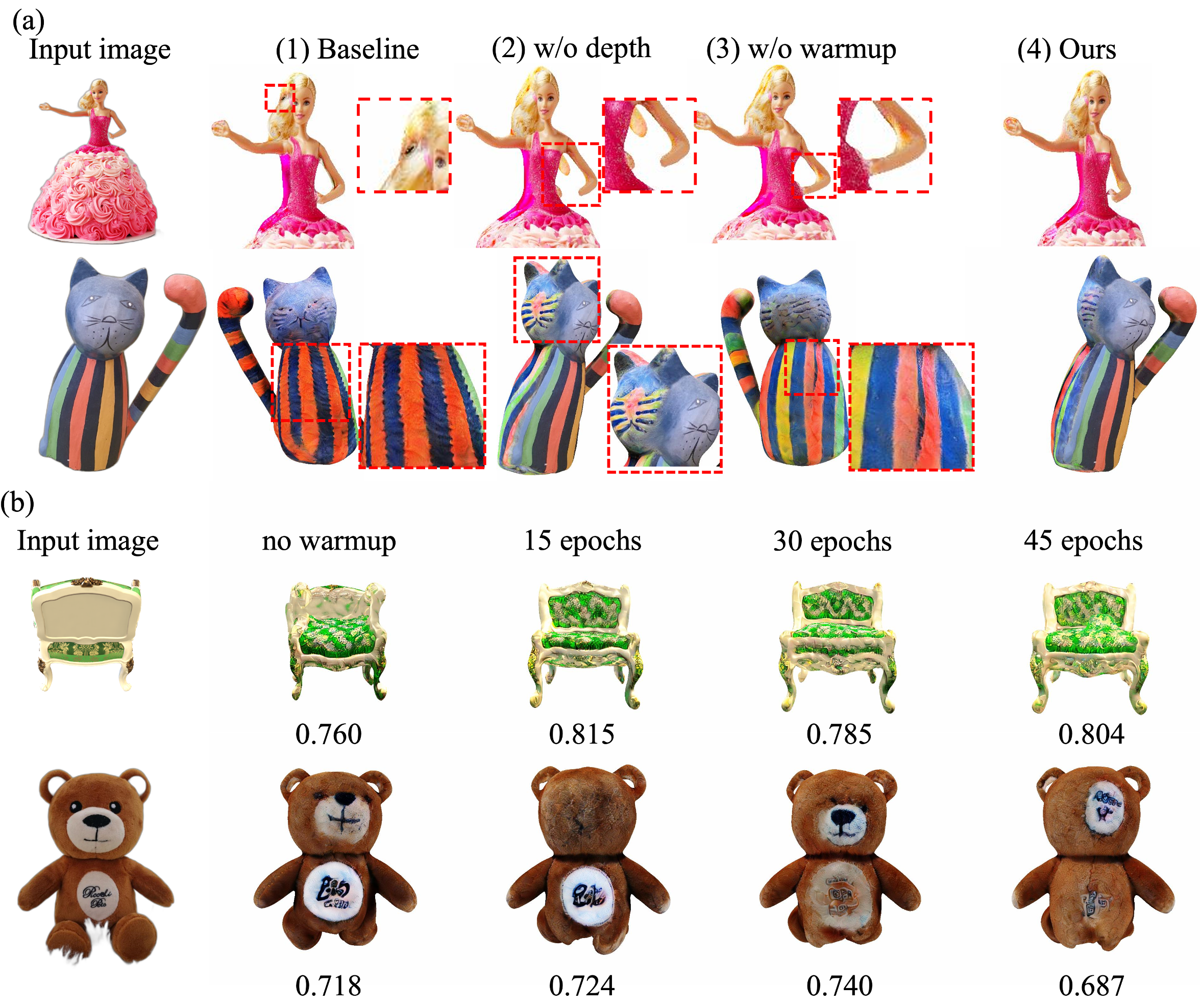}
    \end{adjustbox}
    \caption{(a) Ablation Study. (1) denotes the baseline that does not utilize IP-Adapter. (2) IP-Adapter is utilized, however, attached to naive Stable Diffusion model, not ControlNet.  (3) indicates the absence of depth conditioned warmup strategy. (b) The effects of depth conditioned warmup strategy. (up) For n epochs, only $\hat{\epsilon}_{3D}$ is backpropagated to NeRF model. (down) CLIP-similarity results of each setup.}
    \label{fig:ablation}
\end{figure}

\subsection{Qualitative Results} 
In Fig.~\ref{fig:qualitative} and Fig.~\ref{fig:qualitative_2}, we provide a comparison of visualizations of novel view synthesis by DreamFusion~\cite{poole2022dreamfusion}, RealFusion~\cite{melaskyriazi2023realfusion}, LRM~\cite{hong2023lrm}, LGM~\cite{tang2024lgm}, Magic123~\cite{qian2023magic123} (both with textual inversion and without textual inversion), and our method. Ours consistently generates novel views that closely preserve the detailed textual and geometric features of the given input image, whereas other methods often lose high-frequency details. More examples for qualitative experiments can be found on our GitHub repository.

In contrast to alternative methods, by adding control conditions to the input image, our model is possible to generate controllable 3D objects. Fig.~\ref{fig:control} demonstrates our model can incorporate various conditions such as scribble, pose, depth, and canny. Therefore, our method can utilize not only sketch, but also various optional conditions. 

As shown in Fig.~\ref{fig:control_comparison}, we also compare in controllable 3D object generation with previous work, Control3D~\cite{chen2023control3d} that uses text prompt and sketch condition as input. When given a sketch condition, our method outperforms not only adheres to the specified conditions faithfully, but also excels in maintaining 3D consistency. It demonstrates the effectiveness of controllable image prompt score distillation sampling.

\begin{table}[t]
\centering
\caption{Quantitative results on NeRF4, RealFusion15(RF15), GSO, and CO3D datasets.}
\label{table:1}
\begin{tabular}{c|c|r|r|rr|r} 
\hline
\multirow{3}{*}{Dataset} & \multirow{3}{*}{Metrics} & \multicolumn{1}{c|}{\multirow{3}{*}{LRM~\cite{herlocker2004evaluating}}} & \multicolumn{1}{c|}{\multirow{3}{*}{LGM~\cite{tang2024lgm}}} & \multicolumn{2}{c|}{Magic123~\cite{qian2023magic123}}                     & \multicolumn{1}{c}{\multirow{3}{*}{Ours}}  \\
                         &                          & \multicolumn{1}{c|}{}                     & \multicolumn{1}{c|}{}                     & \multicolumn{2}{c|}{textual inversion}            & \multicolumn{1}{c}{}                       \\
                         &                          & \multicolumn{1}{c|}{}                     & \multicolumn{1}{c|}{}                     & \multicolumn{1}{c}{w/o} & \multicolumn{1}{c|}{w/} & \multicolumn{1}{c}{}                       \\ 
\hline
\multirow{3}{*}{RF15~\cite{melaskyriazi2023realfusion}~}   & CLIP-similarity$\uparrow$          & 0.705                                     & 0.529                                     & 0.827                   & 0.828                   & \textbf{0.850}                             \\
                         & A-LPIPS\textsubscript{VGG}$\downarrow$                  & 0.0071                                    & 0.0109                                    & 0.0790                  & 0.0770                  & 0.0342                                     \\
                         & A-LPIPS\textsubscript{Alex}$\downarrow$                  & \textbf{0.0036}                           & 0.0069                                    & 0.0559                  & 0.0536                  & 0.0213                                     \\ 
\hline
\multirow{3}{*}{NeRF4~\cite{mildenhall2020nerf}~}  & CLIP-similarity$\downarrow$          & 0.615                                     & 0.480                                     & 0.774                   & 0.747                   & \textbf{0.782}                             \\
                         & A-LPIPS\textsubscript{VGG}$\downarrow$                  & \textbf{0.0083}                           & 0.0115                                    & 0.0683                  & 0.0699                  & 0.0335                                     \\
                         & A-LPIPS\textsubscript{Alex}$\downarrow$                  & \textbf{0.0055}                           & 0.0097                                    & 0.0537                  & 0.0519                  & 0.0213                                     \\ 
\hline
\multirow{3}{*}{GSO~\cite{downs2022googlescannedobjectshighquality}}     & CLIP-similarity$\uparrow$          & 0.710                                     & 0.564                                     & 0.737                   & \textbf{0.763}          & 0.761                                      \\
                         & A-LPIPS\textsubscript{VGG}$\downarrow$                  & 0.0075                                    & 0.0056                                    & 0.0165                  & 0.0143                  & 0.0121                                     \\
                         & A-LPIPS\textsubscript{Alex}$\downarrow$                  & \textbf{0.0044}                           & 0.0055                                    & 0.0110                  & 0.0088                  & 0.0077                                     \\ 
\hline
\multirow{3}{*}{CO3D~\cite{reizenstein21co3d}}    & CLIP-similarity$\uparrow$          & 0.630                                     & 0.546                                     & 0.702                   & 0.670                   & \textbf{0.730}                             \\
                         & A-LPIPS\textsubscript{VGG}$\downarrow$                  & 0.0072                                    & \textbf{0.0055}                           & 0.0137                  & 0.0121                  & 0.0110                                     \\
                         & A-LPIPS\textsubscript{Alex}$\downarrow$                  & \textbf{0.0041}                           & 0.0049                                    & 0.0101                  & 0.0092                  & 0.0082                                     \\ 
\hline
\multicolumn{2}{c|}{Train time(m)$\downarrow$}                  & \(\sim 1\)                        & \(\sim 1\)                       & \(\sim 30\)     & \(\sim 120\)   & \(\sim 30\)                        \\
\hline
\end{tabular}
\end{table}

\subsection{Discussion} 
In Fig.~\ref{fig:ablation} (a), we conduct an ablation study on the baseline of our method. (1) Without IP-Adapter, undesirable artifacts are produced as textual inversion trained with a single image fails to capture input image's concepts or styles in detail. (2) Without depth condition (refer to Sec. 3.1), 3D model fails to maintain 3D consistency. (3) Without depth conditioned warmup strategy (refer to Sec. 3.2), 3D consistency is also adversely affected because predicted depth is unstable in early epochs.

We examine the effects of a depth conditioned warmup strategy in our method. As depicted in Fig.~\ref{fig:ablation} (b), the the edge case where a warmup stage is absent, the fidelity of 3D object is compromised. As shown in Table \ref{table:2}, we investigated optimal epochs \( n \) for the warmup strategy. Backpropagating only $\hat{\epsilon}_{3D}$ to the NeRF model achieves the optimal value of CLIP-similarity when n is set to 15, based on our experimental results. Fig.~\ref{fig:ablation} (b) also provides detailed visualization results from front view to most extreme case of the rear view of the input image's camera pose.

\begin{figure}[!t]
    \begin{adjustbox}{width=1.3\linewidth, center}
        \includegraphics{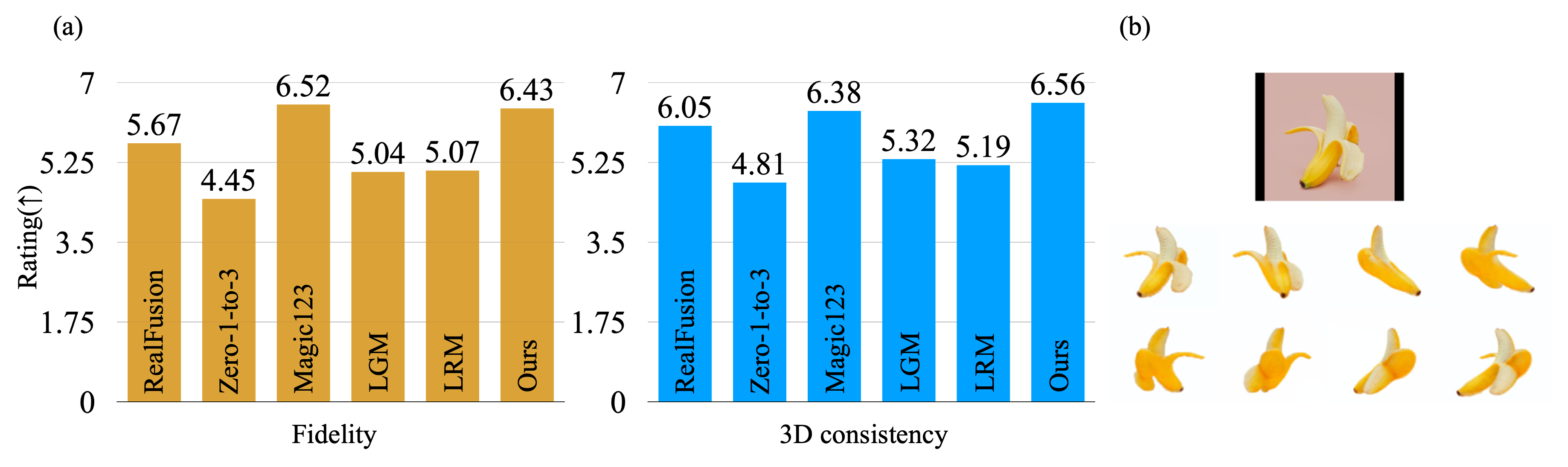}
    \end{adjustbox}
    \caption{(a) User study evaluation of the generated 3D objects. Participants are asked to evaluate two questions regarding fidelity(left) and 3D consistency(right). The rating scale is 1-10. (b) User study form. We show participants an input image(top) and 8 views(including reference view) of the 3D object trained on the input image.}
    \label{fig:user_study}
\end{figure}

\subsection{User Study}
While existing metrics for evaluating text-to-3D models are available~\cite{he2023t3bench,wu2024gpt4vision}, the reliability of using CLIP-similarity and A-LPIPS for evaluating image-to-3D models remains questionable~\cite{katzir2023noisefree}. Therefore, we evaluate the alignment of generated 3D results with human preferences through a user study. In user study, participants are shown the input images alongside eight rendered views of 3D objects, with evenly spaced azimuths (see Fig. \ref{fig:user_study}). They are then asked to respond to two questions: (i) Fidelity, (ii) 3D consistency. Fidelity evaluates how closely generated 3D objects visually match the object in the input image. 3D consistency, regardless of the input image, assesses how naturally results from any camera viewpoint and their freedom from Janus problem~\cite{armandpour2023re}. Responses are rated on a scale from 1 to 10, with higher scores indicating better performance.

We compare our method with alternative methods such as RealFusion, Zero-1-to-3, Magic123, LRM, and LGM. As depicted in Fig.~\ref{fig:user_study}, our method shows matched or better performance against the existing alternatives in terms of both fidelity and 3D consistency. These results validate the effectiveness of our method.

\begin{table}[!t]
\centering
\caption{Variation of CLIP-similarity with varying warmup epochs (n) on the depth-conditioned warmup strategy.}
\label{table:2}
\begin{tblr}{
  cells = {c},
  cell{2}{2} = {r},
  cell{2}{3} = {r},
  cell{2}{4} = {r},
  cell{2}{5} = {r},
  hlines,
  vline{2-5} = {-}{},
}
warmup epochs(n) & 0~    & 15    & 30    & 45    \\
CLIP-similarity  & 0.830 & \textbf{0.834} & 0.829 & 0.784 
\end{tblr}
\end{table}

\section{Conclusion}
In this paper, we proposed a novel approach for the image-to-3D generation task. While existing methods employ text-guided priors and textual inversion to derive text prompt, our method is directly utilizes a single image prompt. ours allows for the incorporating optional conditions such as depth, pose, and text to facilitate more controllable generation of 3D object. Additionally, we propose depth conditioned warmup strategy to enhance 3D consistency. In the public benchmark, our method shows comparable performance to the existing state-of-the-art alternatives.

\subsubsection{Acknowledgements} 
{\small This research was supported by Institute of Information \& Communications Technology Planning \& Evaluation(IITP) grant (No.RS-2022-00167194; Trustworthy AI on Mission Critical Systems, IITP-2024-RS-2024-00357879; AI-based Biosignal Fusion and Generation Technology for Intelligent Personalized Chronic Disease Management, IITP-2024-RS-2024-00417958; Global Research Support Program in the Digital Field) funded by the Korea government(MSIT).}

\bibliographystyle{splncs04}
\bibliography{refs}

@misc{gal2022image,
      title={An Image is Worth One Word: Personalizing Text-to-Image Generation using Textual Inversion}, 
      author={Rinon Gal and Yuval Alaluf and Yuval Atzmon and Or Patashnik and Amit H. Bermano and Gal Chechik and Daniel Cohen-Or},
      year={2022},
      eprint={2208.01618},
      archivePrefix={arXiv},
      primaryClass={cs.CV}
}

@article{poole2022dreamfusion,
  author = {Poole, Ben and Jain, Ajay and Barron, Jonathan T. and Mildenhall, Ben},
  title = {DreamFusion: Text-to-3D using 2D Diffusion},
  journal = {arXiv},
  year = {2022},
}

@inproceedings{lin2023magic3d,
  title={Magic3D: High-Resolution Text-to-3D Content Creation},
  author={Lin, Chen-Hsuan and Gao, Jun and Tang, Luming and Takikawa, Towaki and Zeng, Xiaohui and Huang, Xun and Kreis, Karsten and Fidler, Sanja and Liu, Ming-Yu and Lin, Tsung-Yi},
  booktitle={IEEE Conference on Computer Vision and Pattern Recognition ({CVPR})},
  year={2023}
}

@InProceedings{Chen_2023_ICCV,
    author    = {Chen, Rui and Chen, Yongwei and Jiao, Ningxin and Jia, Kui},
    title     = {Fantasia3D: Disentangling Geometry and Appearance for High-quality Text-to-3D Content Creation},
    booktitle = {Proceedings of the IEEE/CVF International Conference on Computer Vision (ICCV)},
    month     = {October},
    year      = {2023}
}

@article{ye2023ip-adapter,
  title={IP-Adapter: Text Compatible Image Prompt Adapter for Text-to-Image Diffusion Models},
  author={Ye, Hu and Zhang, Jun and Liu, Sibo and Han, Xiao and Yang, Wei},
  booktitle={arXiv preprint arxiv:2308.06721},
  year={2023}
}

@inproceedings{mildenhall2020nerf,
  title={NeRF: Representing Scenes as Neural Radiance Fields for View Synthesis},
  author={Ben Mildenhall and Pratul P. Srinivasan and Matthew Tancik and Jonathan T. Barron and Ravi Ramamoorthi and Ren Ng},
  year={2020},
  booktitle={ECCV},
}

@article{mueller2022instant,
    author = {Thomas M\"uller and Alex Evans and Christoph Schied and Alexander Keller},
    title = {Instant Neural Graphics Primitives with a Multiresolution Hash Encoding},
    journal = {ACM Trans. Graph.},
    issue_date = {July 2022},
    volume = {41},
    number = {4},
    month = jul,
    year = {2022},
    pages = {102:1--102:15},
    articleno = {102},
    numpages = {15},
    url = {https://doi.org/10.1145/3528223.3530127},
    doi = {10.1145/3528223.3530127},
    publisher = {ACM},
    address = {New York, NY, USA},
}

@inproceedings{shen2021dmtet,
        title = {Deep Marching Tetrahedra: a Hybrid Representation for High-Resolution 3D Shape Synthesis},
        author = {Tianchang Shen and Jun Gao and Kangxue Yin and Ming-Yu Liu and Sanja Fidler},
        year = {2021},
        booktitle = {Advances in Neural Information Processing Systems (NeurIPS)}
        }

@InProceedings{Rombach_2022_CVPR,
    author    = {Rombach, Robin and Blattmann, Andreas and Lorenz, Dominik and Esser, Patrick and Ommer, Bj\"orn},
    title     = {High-Resolution Image Synthesis With Latent Diffusion Models},
    booktitle = {Proceedings of the IEEE/CVF Conference on Computer Vision and Pattern Recognition (CVPR)},
    month     = {June},
    year      = {2022},
    pages     = {10684-10695}
}

@misc{zhang2023adding,
  title={Adding Conditional Control to Text-to-Image Diffusion Models}, 
  author={Lvmin Zhang and Maneesh Agrawala},
  year={2023},
  eprint={2302.05543},
  archivePrefix={arXiv},
  primaryClass={cs.CV}
}

@inproceedings{melaskyriazi2023realfusion,
  author = {Melas-Kyriazi, Luke and Rupprecht, Christian and Laina, Iro and Vedaldi, Andrea},
  title = {RealFusion: 360 Reconstruction of Any Object from a Single Image},
  booktitle={CVPR},
  year = {2023},
  url = {https://arxiv.org/abs/2302.10663},
}

@article{qian2023magic123,
  title={Magic123: One Image to High-Quality 3D Object Generation Using Both 2D and 3D Diffusion Priors},
  author={Qian, Guocheng and Mai, Jinjie and Hamdi, Abdullah and Ren, Jian and Siarohin, Aliaksandr and Li, Bing and Lee, Hsin-Ying and Skorokhodov, Ivan and Wonka, Peter and Tulyakov, Sergey and Ghanem, Bernard},
  journal={arXiv preprint arXiv:2306.17843},
  year={2023}
}

@misc{hong2023debiasing,
      title={Debiasing Scores and Prompts of 2D Diffusion for Robust Text-to-3D Generation}, 
      author={Susung Hong and Donghoon Ahn and Seungryong Kim},
      year={2023},
      eprint={2303.15413},
      archivePrefix={arXiv},
      primaryClass={cs.CV}
}

@misc{simonyan2015deep,
      title={Very Deep Convolutional Networks for Large-Scale Image Recognition}, 
      author={Karen Simonyan and Andrew Zisserman},
      year={2015},
      eprint={1409.1556},
      archivePrefix={arXiv},
      primaryClass={cs.CV}
}

@article{krizhevsky2012imagenet,
  title={Imagenet classification with deep convolutional neural networks},
  author={Krizhevsky, Alex and Sutskever, Ilya and Hinton, Geoffrey E},
  journal={Advances in neural information processing systems},
  volume={25},
  year={2012}
}

@misc{zhang2018unreasonable,
      title={The Unreasonable Effectiveness of Deep Features as a Perceptual Metric}, 
      author={Richard Zhang and Phillip Isola and Alexei A. Efros and Eli Shechtman and Oliver Wang},
      year={2018},
      eprint={1801.03924},
      archivePrefix={arXiv},
      primaryClass={cs.CV}
}

@article{armandpour2023re,
        title={Re-imagine the Negative Prompt Algorithm: Transform 2D Diffusion into 3D, alleviate Janus problem and Beyond},
        author={Armandpour, Mohammadreza and Zheng, Huangjie and Sadeghian, Ali and Sadeghian, Amir and Zhou, Mingyuan},
        journal={arXiv preprint arXiv:2304.04968},
        year={2023}
      }

@misc{liu2023zero1to3,
      title={Zero-1-to-3: Zero-shot One Image to 3D Object}, 
      author={Ruoshi Liu and Rundi Wu and Basile Van Hoorick and Pavel Tokmakov and Sergey Zakharov and Carl Vondrick},
      year={2023},
      eprint={2303.11328},
      archivePrefix={arXiv},
      primaryClass={cs.CV}
}

@misc{ho2020denoising,
      title={Denoising Diffusion Probabilistic Models}, 
      author={Jonathan Ho and Ajay Jain and Pieter Abbeel},
      year={2020},
      eprint={2006.11239},
      archivePrefix={arXiv},
      primaryClass={cs.LG}
}

@misc{ramesh2022hierarchical,
      title={Hierarchical Text-Conditional Image Generation with CLIP Latents}, 
      author={Aditya Ramesh and Prafulla Dhariwal and Alex Nichol and Casey Chu and Mark Chen},
      year={2022},
      eprint={2204.06125},
      archivePrefix={arXiv},
      primaryClass={cs.CV}
}

@article{schuhmann2022laion,
  title={Laion-5b: An open large-scale dataset for training next generation image-text models},
  author={Schuhmann, Christoph and Beaumont, Romain and Vencu, Richard and Gordon, Cade and Wightman, Ross and Cherti, Mehdi and Coombes, Theo and Katta, Aarush and Mullis, Clayton and Wortsman, Mitchell and others},
  journal={Advances in Neural Information Processing Systems},
  volume={35},
  pages={25278--25294},
  year={2022}
}

@misc{radford2021learning,
      title={Learning Transferable Visual Models From Natural Language Supervision}, 
      author={Alec Radford and Jong Wook Kim and Chris Hallacy and Aditya Ramesh and Gabriel Goh and Sandhini Agarwal and Girish Sastry and Amanda Askell and Pamela Mishkin and Jack Clark and Gretchen Krueger and Ilya Sutskever},
      year={2021},
      eprint={2103.00020},
      archivePrefix={arXiv},
      primaryClass={cs.CV}
}

@misc{chen2023control3d,
      title={Control3D: Towards Controllable Text-to-3D Generation}, 
      author={Yang Chen and Yingwei Pan and Yehao Li and Ting Yao and Tao Mei},
      year={2023},
      eprint={2311.05461},
      archivePrefix={arXiv},
      primaryClass={cs.CV}
}

@inproceedings{lee2022bridging,
  title={Bridging the domain gap towards generalization in automatic colorization},
  author={Lee, Hyejin and Kim, Daehee and Lee, Daeun and Kim, Jinkyu and Lee, Jaekoo},
  booktitle={European Conference on Computer Vision},
  pages={527--543},
  year={2022},
  organization={Springer}
}

@InProceedings{Zhu_2023_CVPR_tryondiffusion,
  author={Zhu, Luyang and Yang, Dawei and Zhu, Tyler and Reda, Fitsum and Chan, William and Saharia, Chitwan and Norouzi, Mohammad and Kemelmacher-Shlizerman, Ira},
  title={TryOnDiffusion: A Tale of Two UNets},
  booktitle = {Proceedings of the IEEE/CVF Conference on Computer Vision and Pattern Recognition (CVPR)},
  month = {June},
  year={2023},
  pages = {4606-4615}
}

@article{ranftl2020towards,
  title={Towards robust monocular depth estimation: Mixing datasets for zero-shot cross-dataset transfer},
  author={Ranftl, Ren{\'e} and Lasinger, Katrin and Hafner, David and Schindler, Konrad and Koltun, Vladlen},
  journal={IEEE transactions on pattern analysis and machine intelligence},
  volume={44},
  number={3},
  pages={1623--1637},
  year={2020},
  publisher={IEEE}
}

@inproceedings{Mohammad_Khalid_2022, series={SA ’22},
   title={CLIP-Mesh: Generating textured meshes from text using pretrained image-text models},
   url={http://dx.doi.org/10.1145/3550469.3555392},
   DOI={10.1145/3550469.3555392},
   booktitle={SIGGRAPH Asia 2022 Conference Papers},
   publisher={ACM},
   author={Mohammad Khalid, Nasir and Xie, Tianhao and Belilovsky, Eugene and Popa, Tiberiu},
   year={2022},
   month=nov, collection={SA ’22} }

@misc{sora,
    title={Sora: Creating video from text.},
    author={OpenAI},
    url={https://openai.com/sora},
    year={2024}
}

@misc{zabari2023diffusing,
      title={Diffusing Colors: Image Colorization with Text Guided Diffusion}, 
      author={Nir Zabari and Aharon Azulay and Alexey Gorkor and Tavi Halperin and Ohad Fried},
      year={2023},
      eprint={2312.04145},
      archivePrefix={arXiv},
      primaryClass={cs.CV}
}

@misc{katzir2023noisefree,
      title={Noise-Free Score Distillation}, 
      author={Oren Katzir and Or Patashnik and Daniel Cohen-Or and Dani Lischinski},
      year={2023},
      eprint={2310.17590},
      archivePrefix={arXiv},
      primaryClass={cs.CV}
}

@misc{he2023t3bench,
      title={T$^3$Bench: Benchmarking Current Progress in Text-to-3D Generation}, 
      author={Yuze He and Yushi Bai and Matthieu Lin and Wang Zhao and Yubin Hu and Jenny Sheng and Ran Yi and Juanzi Li and Yong-Jin Liu},
      year={2023},
      eprint={2310.02977},
      archivePrefix={arXiv},
      primaryClass={cs.CV}
}

@misc{wu2024gpt4vision,
      title={GPT-4V(ision) is a Human-Aligned Evaluator for Text-to-3D Generation}, 
      author={Tong Wu and Guandao Yang and Zhibing Li and Kai Zhang and Ziwei Liu and Leonidas Guibas and Dahua Lin and Gordon Wetzstein},
      year={2024},
      eprint={2401.04092},
      archivePrefix={arXiv},
      primaryClass={cs.CV}
}

@misc{meng2022sdedit,
      title={SDEdit: Guided Image Synthesis and Editing with Stochastic Differential Equations}, 
      author={Chenlin Meng and Yutong He and Yang Song and Jiaming Song and Jiajun Wu and Jun-Yan Zhu and Stefano Ermon},
      year={2022},
      eprint={2108.01073},
      archivePrefix={arXiv},
      primaryClass={cs.CV}
}

@misc{lugmayr2022repaint,
      title={RePaint: Inpainting using Denoising Diffusion Probabilistic Models}, 
      author={Andreas Lugmayr and Martin Danelljan and Andres Romero and Fisher Yu and Radu Timofte and Luc Van Gool},
      year={2022},
      eprint={2201.09865},
      archivePrefix={arXiv},
      primaryClass={cs.CV}
}

@misc{sinha2021d2c,
      title={D2C: Diffusion-Denoising Models for Few-shot Conditional Generation}, 
      author={Abhishek Sinha and Jiaming Song and Chenlin Meng and Stefano Ermon},
      year={2021},
      eprint={2106.06819},
      archivePrefix={arXiv},
      primaryClass={cs.LG}
}

@misc{saharia2022palette,
      title={Palette: Image-to-Image Diffusion Models}, 
      author={Chitwan Saharia and William Chan and Huiwen Chang and Chris A. Lee and Jonathan Ho and Tim Salimans and David J. Fleet and Mohammad Norouzi},
      year={2022},
      eprint={2111.05826},
      archivePrefix={arXiv},
      primaryClass={cs.CV}
}

@misc{li2021srdiff,
      title={SRDiff: Single Image Super-Resolution with Diffusion Probabilistic Models}, 
      author={Haoying Li and Yifan Yang and Meng Chang and Huajun Feng and Zhihai Xu and Qi Li and Yueting Chen},
      year={2021},
      eprint={2104.14951},
      archivePrefix={arXiv},
      primaryClass={cs.CV}
}

@misc{saharia2021image,
      title={Image Super-Resolution via Iterative Refinement}, 
      author={Chitwan Saharia and Jonathan Ho and William Chan and Tim Salimans and David J. Fleet and Mohammad Norouzi},
      year={2021},
      eprint={2104.07636},
      archivePrefix={arXiv},
      primaryClass={eess.IV}
}

@misc{singer2022makeavideo,
      title={Make-A-Video: Text-to-Video Generation without Text-Video Data}, 
      author={Uriel Singer and Adam Polyak and Thomas Hayes and Xi Yin and Jie An and Songyang Zhang and Qiyuan Hu and Harry Yang and Oron Ashual and Oran Gafni and Devi Parikh and Sonal Gupta and Yaniv Taigman},
      year={2022},
      eprint={2209.14792},
      archivePrefix={arXiv},
      primaryClass={cs.CV}
}

@misc{chen2023text2tex,
      title={Text2Tex: Text-driven Texture Synthesis via Diffusion Models}, 
      author={Dave Zhenyu Chen and Yawar Siddiqui and Hsin-Ying Lee and Sergey Tulyakov and Matthias Nießner},
      year={2023},
      eprint={2303.11396},
      archivePrefix={arXiv},
      primaryClass={cs.CV}
}

@misc{rombach2022highresolution,
      title={High-Resolution Image Synthesis with Latent Diffusion Models}, 
      author={Robin Rombach and Andreas Blattmann and Dominik Lorenz and Patrick Esser and Björn Ommer},
      year={2022},
      eprint={2112.10752},
      archivePrefix={arXiv},
      primaryClass={cs.CV}
}

@inproceedings{xu2023neurallift,
  title={Neurallift-360: Lifting an in-the-wild 2d photo to a 3d object with 360deg views},
  author={Xu, Dejia and Jiang, Yifan and Wang, Peihao and Fan, Zhiwen and Wang, Yi and Wang, Zhangyang},
  booktitle={Proceedings of the IEEE/CVF Conference on Computer Vision and Pattern Recognition},
  pages={4479--4489},
  year={2023}
}

@article{Ranftl2021,
	author    = {Ren\'{e} Ranftl and Alexey Bochkovskiy and Vladlen Koltun},
	title     = {Vision Transformers for Dense Prediction},
	journal   = {ArXiv preprint},
	year      = {2021},
}

@inproceedings{reizenstein21co3d,
    Author = {Reizenstein, Jeremy and Shapovalov, Roman and Henzler, Philipp and Sbordone, Luca and Labatut, Patrick and Novotny, David},
    Booktitle = {International Conference on Computer Vision},
    Title = {Common Objects in 3D: Large-Scale Learning and Evaluation of Real-life 3D Category Reconstruction},
    year = {2021},
}

@misc{downs2022googlescannedobjectshighquality,
      title={Google Scanned Objects: A High-Quality Dataset of 3D Scanned Household Items}, 
      author={Laura Downs and Anthony Francis and Nate Koenig and Brandon Kinman and Ryan Hickman and Krista Reymann and Thomas B. McHugh and Vincent Vanhoucke},
      year={2022},
      eprint={2204.11918},
      archivePrefix={arXiv},
      primaryClass={cs.RO},
      url={https://arxiv.org/abs/2204.11918}, 
}

@article{hong2023lrm,
  title={Lrm: Large reconstruction model for single image to 3d},
  author={Hong, Yicong and Zhang, Kai and Gu, Jiuxiang and Bi, Sai and Zhou, Yang and Liu, Difan and Liu, Feng and Sunkavalli, Kalyan and Bui, Trung and Tan, Hao},
  journal={arXiv preprint arXiv:2311.04400},
  year={2023}
}

@article{herlocker2004evaluating,
  title={Evaluating collaborative filtering recommender systems},
  author={Herlocker, Jonathan L and Konstan, Joseph A and Terveen, Loren G and Riedl, John T},
  journal={ACM Transactions on Information Systems (TOIS)},
  volume={22},
  number={1},
  pages={5--53},
  year={2004},
  publisher={ACM New York, NY, USA}
}

@article{tang2024lgm,
  title={LGM: Large Multi-View Gaussian Model for High-Resolution 3D Content Creation},
  author={Tang, Jiaxiang and Chen, Zhaoxi and Chen, Xiaokang and Wang, Tengfei and Zeng, Gang and Liu, Ziwei},
  journal={arXiv preprint arXiv:2402.05054},
  year={2024}
}

@misc{lin2015microsoftcococommonobjects,
      title={Microsoft COCO: Common Objects in Context}, 
      author={Tsung-Yi Lin and Michael Maire and Serge Belongie and Lubomir Bourdev and Ross Girshick and James Hays and Pietro Perona and Deva Ramanan and C. Lawrence Zitnick and Piotr Dollár},
      year={2015},
      eprint={1405.0312},
      archivePrefix={arXiv},
      primaryClass={cs.CV},
      url={https://arxiv.org/abs/1405.0312}, 
}

\end{document}